\documentclass[runningheads]{llncs}
\usepackage{fontspec}
\usepackage[T1]{fontenc}
\usepackage{graphicx}
\usepackage{hyperref}
\usepackage{booktabs}
\usepackage{siunitx}
\usepackage{multirow}
\usepackage[most]{tcolorbox}
\usepackage[table]{xcolor}
\tcbset{colback=yellow!10!white, colframe=yellow!50!black, sharp corners}
\usepackage{colortbl}
\usepackage{tabularx}
\usepackage{array}
\newcolumntype{Y}{>{\centering\arraybackslash}X}
\usepackage{tikz}
\usetikzlibrary{arrows.meta,positioning,shapes.geometric}
\usepackage{subcaption}

\newif\ifFinale
\Finaletrue   % Set to true
\newcommand{\finalonly}[1]{%
  \ifFinale#1\fi
}
\usepackage{xcolor}
\usepackage[normalem]{ulem} % for strikeout

\definecolor{DIFdelcolor}{RGB}{180,0,0}   % dark red
\definecolor{DIFaddcolor}{RGB}{0,120,0}   % dark green

\newcommand{\DIFdel}[1]{%
  {\color{DIFdelcolor}\sout{#1}}%
}

\newcommand{\DIFadd}[1]{%
  {\color{DIFaddcolor}#1}%
}

\newcommand{\DIFdelbegin}{}
\newcommand{\DIFdelend}{}
\newcommand{\DIFaddbegin}{}
\newcommand{\DIFaddend}{}
\usepackage{polyglossia}
\newfontfamily\greekfont[Path=./, Extension=.ttf]{GFSDidot}
\setmainlanguage{english}
\setotherlanguage{greek}

\begin{document}

\title{Structure-Aware Text Recognition\\for Ancient Greek Critical Editions}

\finalonly{
\author{Nicolas Angleraud\inst{1}\orcidID{0009-0008-9076-7630} \and Antonia Karamolegkou\inst{1}\orcidID{0000-0002-6458-0986} \and Benoît Sagot\inst{1}\orcidID{ 0000-0002-0107-8526} \and Thibault Clérice\inst{1}\orcidID{
0000-0003-1852-9204}}
\institute{Inria, Paris, France 
\email{\{firstname.familyname\}@inria.fr}}}
\maketitle              % typeset the header of the contribution
\begin{abstract}
Recent advances in visual language models (VLMs) have transformed end-to-end document understanding. However, their ability to interpret the complex layout semantics of historical scholarly texts remains limited. This paper investigates structure-aware text recognition for Ancient Greek critical editions, which have dense reference hierarchies and extensive marginal annotations. We introduce two novel resources: (i) a large-scale synthetic corpus of 185,000 page images generated from TEI/XML sources with controlled typographic and layout variation, and (ii) a curated benchmark of real scanned editions spanning more than a century of editorial and typographic practices. Using these datasets, we evaluate three state-of-the-art VLMs under both zero-shot and fine-tuning regimes. Our experiments reveal substantial limitations in current VLM architectures when confronted with highly structured historical documents. In zero-shot settings, most models underperform compared to established off-the-shelf traditional software. Nevertheless, the Qwen3VL-8B model achieves state-of-the-art performance, reaching a median Character Error Rate of 1.0\% on real scans. These results highlight both the current shortcomings and the future potential of VLMs for structure-aware recognition of complex scholarly documents.
\end{abstract}

\section{Introduction}

Traditional OCR pipelines decompose the task into sequential stages (layout analysis, segmentation, recognition), which are brittle under heterogeneous or complex layouts and noisy scans, and early errors propagate to downstream processing and retrieval. Recent vision–language models (VLMs)---Qwen2.5-VL~\cite{bai2025qwen3vltechnicalreport}, DeepSeek-OCR-2~\cite{wei2026deepseekocr2visualcausal}, and LightOnOCR~\cite{lightOCR}---address this limitation by performing end-to-end recognition, jointly modeling visual input and text generation. While these models achieve strong results on modern benchmarks~\cite{li2025omnibenchfutureuniversalomnilanguage,kargaran2026glotocrbenchocrmodels}, existing evaluations primarily measure plain-text accuracy and do not isolate failures in document structure.
This limitation becomes critical for historical scholarly documents, where layout encodes essential semantic information. In particular, Ancient Greek critical editions combine hierarchical organization, marginalia, and dense reference systems that are central to scholarly use.

\begin{figure}[t]
    \centering
    \includegraphics[width=\linewidth]{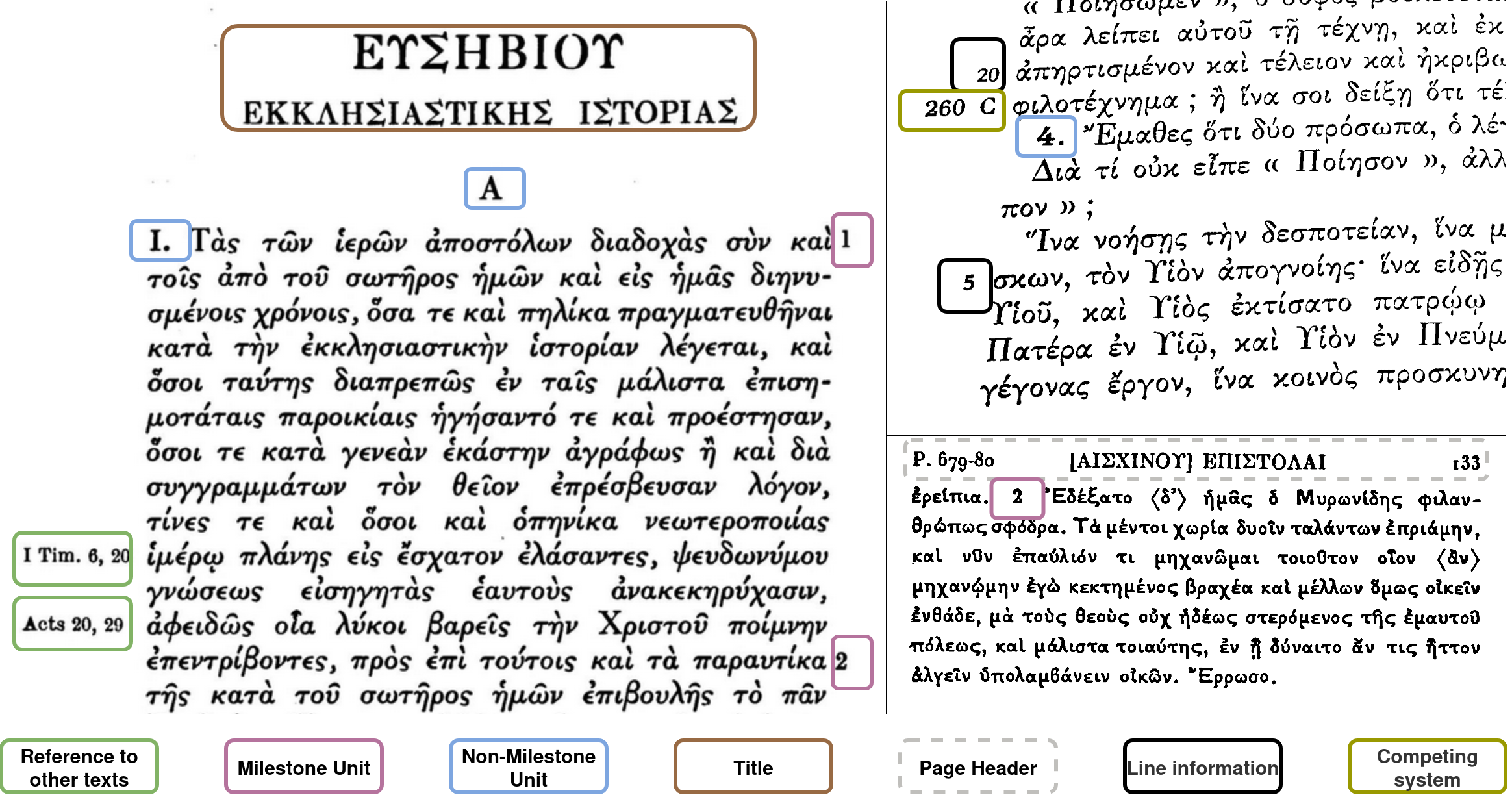}
    \caption{Example of the richness of the layout in three different pages.}
    \label{fig:layoutSemantics}
\end{figure}

Despite prior OCR work on Ancient Greek~\cite{sommerschield-etal-2023-machine,Evaggelatos}, large-scale computational access remains constrained by two bottlenecks. First, open, machine-actionable corpora remain scarce. Much of the textual heritage is accessible only through licensed resources such as the \emph{Thesaurus Linguae Graecae} (TLG), which contains over 12{,}000 works but restricts redistribution, bulk analysis, and derivative use~\cite{tlg2024}. Open corpora derived from printed critical editions remain smaller and uneven in coverage: major alternatives such as the \emph{Perseus Digital Library}~\cite{crane1988perseus} and the \emph{Thousand Years of Ancient Greek}~\cite{muellner2019free} focus mainly on the first millennium of Greek, while Late Antique and Medieval Greek are only sparsely represented, despite targeted extensions~\cite{stockhausen_pta}.
Second, Ancient Greek critical editions pose intrinsic OCR challenges because of their dense and explicit document structure (Figure~\ref{fig:layoutSemantics}). Beyond polytonic script recognition, they combine section hierarchies, milestone numbering, marginal notes, and critical apparatus entries recording variant readings in the textual tradition. These elements are not peripheral: they form an organizational layer used for citation and navigation and interact directly with OCR performance. 

Digitization, therefore, requires not only transcription but also structure recovery, separation of layout artefacts from textual organization, and reconstruction of content fragmented across pages. In practice, this conversion of OCR output into structured TEI/XML representations is often more time-consuming than raw transcription and remains largely manual due to the lack of openly shared ground truth~\cite{TEI_P5}.
In this work, we study structure-aware OCR for Ancient Greek critical editions using a lightweight markup scheme for reference and organizational elements. We construct image--transcription benchmarks spanning more than $5{,}000$ works and supporting both training and evaluation. Although our focus is Ancient Greek, the challenges we target---dense layout, intertwined structure, and multi-layer annotations---are representative of a broader class of complex document understanding problems. Our contributions are:
\begin{itemize}
\item an open large-scale synthetic corpus with controlled typographic and layout variation (185{,}000 images);
\item a reproducible pipeline for generating structure-aware OCR datasets;
\item an open curated benchmark of real printed editions (450 images);
\item a benchmark analysis of state-of-the-art VLMs in zero-shot and fine-tuned settings, with resulting models below 2\% CER\finalonly{ (\href{https://huggingface.co/CLLG}{https://huggingface.co/CLLG})}.
\end{itemize}

\section{Related Work}

The reference corpus for Ancient Greek literary texts remains the \emph{Thesaurus Linguae Graecae} (TLG)\cite{tlg2024}, which offers unmatched coverage from the Archaic period to Byzantium but is distributed under a restrictive license that prohibits redistribution, bulk processing, and derivative datasets. Despite its central role in Greek philology, the TLG therefore cannot support open OCR training, evaluation, or reproducible experimentation. Open alternatives have emerged over the past decades, most notably the Greek collections of the Perseus Digital Library, which established the open dissemination of TEI-encoded Ancient Greek texts. These resources were later consolidated and extended by the \emph{First 1K Years of Greek} (F1KG) project\cite{muellner2019free}, which provides openly licensed literary texts with linguistic annotation and stable citability, and by the \emph{Patristic Text Archive} (PTA)\cite{stockhausen_pta}, which addresses the under-representation of Late Antique texts in other major open corpora.

Beyond continuous literary texts, other Ancient Greek corpora differ substantially in structure and purpose. Fragmentary corpora focusing on lost works transmitted through quotations do not reflect the layout or continuity of book-length texts\cite{berti2019historical}. Epigraphical\cite{orlandi2014eagle} and papyrological\cite{CaylessRouechéElliottBodard+2010+203+222} corpora provide large quantities of Greek text, but their fragmentary supports, non-linear layouts, and strong editorial intervention make them methodologically distinct from printed critical editions. Overall, although several open Ancient Greek corpora exist~\cite{papadopoulos2022greekocr}, their coverage remains partial compared to the TLG, and none provide OCR-oriented ground truth aligned with the structural complexity of critical editions.

Research on automatic processing of Ancient Greek has likewise treated OCR, layout analysis, and post-correction mostly as separate problems. Early OCR work used adaptable frameworks such as Gamera~\cite{doi:10.3138/mous.14.3-3} to recognize polytonic Greek in historical prints, while user-facing environments such as Lace~\cite{robertson2019optical} emphasized interactive post-correction. More recent work has addressed layout analysis for text-heavy historical commentaries~\cite{najem-meyer_page-layout-analysis_2022}, Ancient Greek text recognition~\cite{10.1145/3476887.3476911}, and LLM-based post-correction~\cite{boros-etal-2024-post,pavlopoulos-etal-2024-challenging}. These studies highlight challenges specific to Ancient Greek, including rich morphology, a long diachronic span from approximately 900~BCE to 1500~CE, and strong geographical and dialectal variation. Taken together, prior work either separates layout from text or relies on post-hoc correction, leaving the joint problem of structurally aware OCR for Ancient Greek critical editions largely unexplored.

\section{Layout Semantics in Editions of Ancient Texts}
\label{sec:structure}

\paragraph{What Is Structure in a Critical Edition.} Classical texts in Latin and Greek have been for centuries organized into hierarchical units comparable to modern chapters and books~\cite{fotheringham2007numbers}, and in some cases since their original publication~\cite[p.~13]{canfora2016conservazione}. Over time, these divisions have been standardized through a process of canonicalization. As a result, systems of reference have become central to scholarly practice, enabling stable citation and comparison across editions. Their use has persisted and extends to digital academic platforms, making their recognition and encoding in digital versions of critical editions essential.

Some structural systems correspond closely to contemporary notions of textual organization: chapters and sections often align with paragraph boundaries and coherent thematic units. Others, however, are rooted in historical printing practices and reflect the material constraints of early modern editions (such as the Stephanus pagination, a citation system based on an edition of Plato's work in 1578). Despite their arbitrariness from a modern perspective, these systems continue to function as authoritative reference frameworks. We distinguish between two types of structural levels: milestone and non-milestone units. A milestone structural level is defined by its capacity to appear within a paragraph rather than only at its beginning, enabling fine-grained internal referencing.
Because different systems respond to different scholarly needs, hybrid structures combining multiple non-milestone levels with a final milestone level are common.

\paragraph{Competing Structures and Marginal Information.}

The long-standing tradition of canonical referencing has produced a wide diversity of layout practices for encoding structural information. These often coexist with alternative organizational schemes based on content rather than inherited pagination. For example, although the \emph{Patrologia Graeca} citation system remains the scholarly standard, the philological quality of its editions is often considered insufficient, leading modern editors to propose new hierarchical structures based on chapters, sections, or thematic units. Nevertheless, due to the authority of traditional references, these systems are typically preserved in parallel, often in the margins. Such dual referencing extends to texts transmitted through a single manuscript or authoritative witness, where references to original folios are also provided (e.g., \textit{fol.~123r}, \textit{123v}). Margins also contain additional types of information. In Christian texts, they frequently include references to biblical passages or other theological authorities. More generally, even when no canonical reference is present, line numbering is often provided (e.g., at intervals of 5, 10, 15, or more rarely 4, 8, 12). This layering of marginal, heading, and in-text annotations produces a visually dense and semantically complex layout, in which only a subset of elements has canonical use. For non-specialists, interpreting these intertwined systems can require extensive contextual reading to reconstruct relationships between structural units, marginal references, and editorial interventions. From a document analysis perspective, this richness is both a source of information and a major source of complexity. From a corpus-building perspective, identifying structuring reference identifiers among this noise is the most time-consuming step, as it directly impacts resource encoding~\cite{clerice2017outils}.

\section{Constructing Structure-Aware Resources}
\label{sec:dataset}

\begin{table}[t]
\centering
\caption{Comparison of structural characteristics across synthetic and real data. The total count of pages, the median number of words per page, and the proportion of pages containing at least one instance of the respective structural element (\texttt{<ref>}, \texttt{<note>}, or \texttt{markdown header} for titles and subtitles.}
\small
\begin{tabular*}{\linewidth}{@{\extracolsep{\fill}}lccccc}
\toprule
\textbf{Corpus} & \textbf{Pages} & \textbf{Words$_{\text{med}}$} & 
\textbf{\%Ref} & \textbf{\%Note} & \textbf{\%Head.} \\
\midrule
Synthetic & 185k & 242 & 78.6\% & 6.1\% & 55.2\% \\
Real      & 460  & 215 & 83.0\% & 32.3\% & 30.7\% \\
\bottomrule
\end{tabular*}
\label{tab:dataset_structure_summary}
\end{table}

We construct two complementary resources: (i)~a large dataset of synthetically rendered page images derived from existing critical editions, designed to introduce controlled typographic and layout variation, and (ii)~a small real-data benchmark of scanned editions for measuring generalization under authentic conditions. 
%In both cases, the underlying textual content is drawn from openly licensed sources, and in the synthetic corpus, only the document renderings are synthetic. 
Both resources share the same annotation target: structured Markdown with block-level headings and inline tags (Appendix~\ref{app:annotation-guidelines}).

\subsection{Rendering Structured Editions}
\label{sec:synthetic}

\begin{figure}[t]
    \centering
    \includegraphics[width=0.9\linewidth]{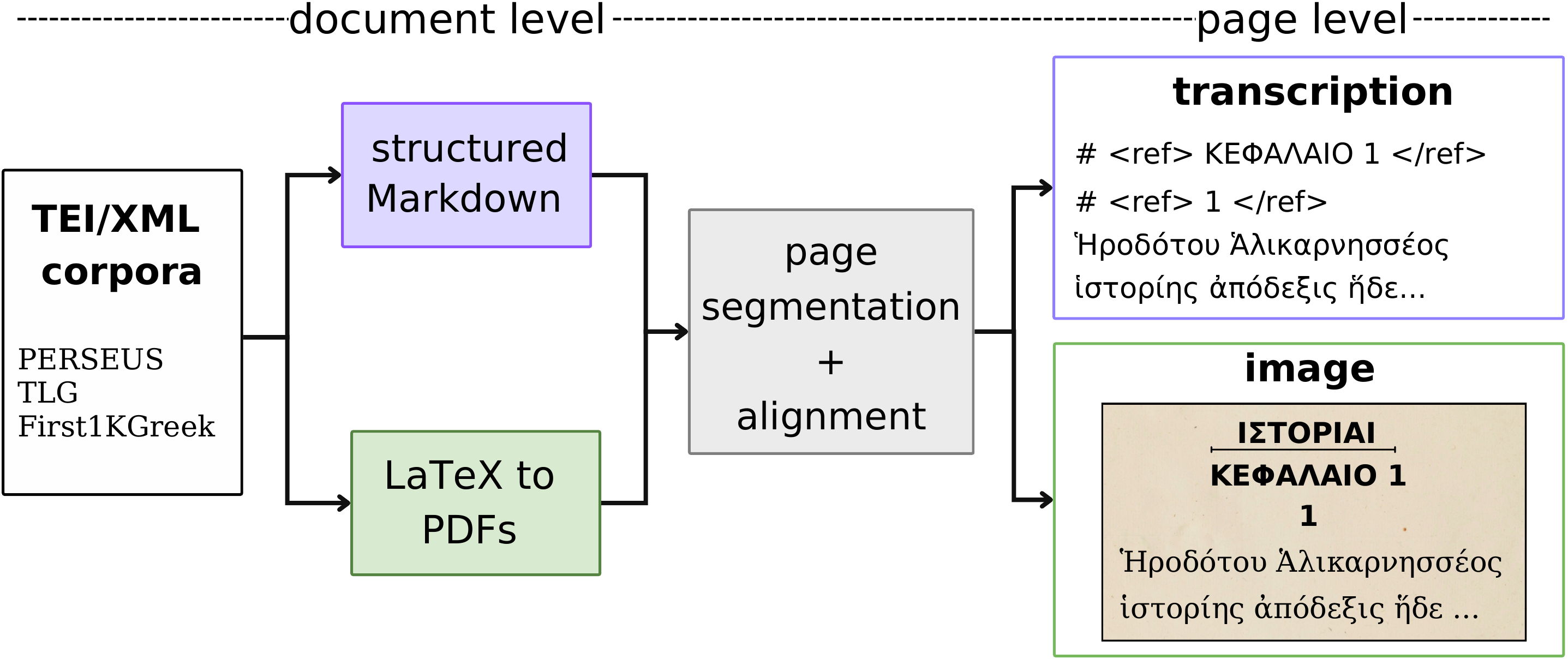}
    \caption{Overview of the synthetic data generation pipeline. TEI/XML corpora are converted into structured Markdown annotations and rendered into PDF page images via LaTeX, producing aligned text--image pairs.}
    \label{fig:framework}
\end{figure}

To enable controlled, large-scale training and evaluation for Ancient Greek OCR under complex scholarly layouts, we generate \emph{synthetic page images} by rendering openly licensed TEI/XML texts into typographically diverse document pages. This preserves the original content while introducing realistic variation in layout and typographical conventions found in printed critical editions. We collect prose texts from TEI/XML corpora of Ancient Greek~\cite{Clerice2026CLLG,greeklit,first1k}, including one Latin corpus~\cite{latinlit}.\footnote{The Latin corpus is included to maintain performance for Latin-script material, which frequently appears in Greek editions (e.g., references, abbreviations, editorial conventions).} Where the same work appears in multiple corpora with different encodings, we retain all versions to increase diversity. The pipeline is summarized in Figure~\ref{fig:framework}.
%We restrict the synthetic corpus to prose texts by filtering the source TEI/XML corpora and excluding works whose primary textual structure is encoded using verse-specific elements %(e.g., \verb|<lg>|, \verb|<l>|).
%This ensures that the rendered documents reflect the paragraph-based layouts typical of prose critical editions.
For each TEI document, we extract a citation-oriented structural description from the TEI tree. This \texttt{citeStructure}\cite{cayless:hal-04262751} specifies the hierarchical units used for citation and navigation (e.g., book, chapter, section) and their nesting. We then use a lightweight pseudo-XML annotation scheme to mark the text and its structure. According to this scheme, the hierarchical units are marked in between \verb|<ref>...</ref>| tags, the editorial notes are marked in \verb|<note>...</note>| tags, and all titles are marked with a \verb|#| following markdown practices. As this paper aims to facilitate the rapid re-encoding of multi-page documents, we also address the issue of paragraphs that are broken across pages. To support their reconstruction, we add a \verb|<tab/>| tag to paragraphs that are indented in the printed layout. This marker enables reliable post-processing and merging of cross-page fragments, ensuring that the synthetic dataset supports both OCR transcription and document-level, structure-aware evaluation.
%can be used not only for OCR transcription but also for the evaluation of structure-aware models operating at the document level.

To introduce controlled visual variation, we surveyed critical editions and, guided by editorial and typographical conventions, constructed a curated set of \LaTeX{} layout configurations and background templates. These modulate page-level and typographic properties, including background color\cite{Chague_Gallicalbum_2023}, page size, column count, font families, numeral styles, and the positioning of titles and structural markers, as well as higher-level parameters such as paper format and content representation, while preserving the underlying text and annotations. For each document, multiple configurations are randomly sampled and applied during rendering, allowing the same source to be realized under diverse typographic conditions without altering its semantic content. The goal is not to reproduce specific documents, but to cover a representative range of plausible configurations grounded in real practices. Prior to rendering, TEI/XML sources are lightly normalized and then transformed via XSLT into a \LaTeX{} source file with layout placeholders and a parallel pseudo-XML target used for supervision and evaluation. This pipeline ensures consistency between visual output and structural annotation while enabling systematic variation in document appearance.

We generate two PDF versions: one in black text and one with color-coded components. The color-coded component allows for identifying information, such as the footer, which is subsequently masked to prevent the introduction of artificial supervision signals.\footnote{Footnotes, specifically variants from other manuscripts, are under a different copyright system in multiple countries, including France and Germany.}Page text is extracted from the color-coded PDF via color filtering to retain only the main-text channel. The full target text is segmented into page-sized chunks using tail matching with a length-based fallback. A page pair is retained only if extracted text and segmented target achieve a Levenshtein similarity of $0.99$, ensuring high-fidelity alignment.

\subsection{Annotating Printed Editions}
\label{subsec:real}

We also construct a compact benchmark of real scanned editions to evaluate zero-shot behavior and transfer after fine-tuning on synthetic data. The dataset comprises 30 author–work pairs spanning approximately 1,800 years of Greek literary production, from Hippocrates (5th c.~BCE) to Nicolas Cabasilas (14th c.~CE). To assess both in-domain and out-of-distribution generalization, half of the authors overlap with the synthetic corpus pool, while the rest are disjoint and increase coverage of post-classical (AD) material. Editions are drawn from diverse editorial traditions: approximately half belong to the \emph{Sources Chrétiennes} series, with additional representation from Oxford/Clarendon, Les Belles Lettres, the \emph{Patrologia Graeca}, the \emph{Griechische Christliche Schriftsteller}, \emph{Zeitschrift für Papyrologie und Epigraphik}, \emph{Revue des études byzantines}, Princeton University Press, and 19th-century scholarly editions (Weidmann, Baillière, Laupp). Publication dates range from 1844 to 2017, covering nearly two centuries of typographic variation.

For each work, we sample 15 pages uniformly at random and manually annotate them following Appendix~\ref{app:annotation-guidelines}. The annotation scheme mirrors the synthetic data and encodes heading patterns and inline reference tags (\verb|<ref>...</ref>|, \verb|<note>...</note>|). Table~\ref{tab:dataset_structure_summary} summarizes key statistics: 83\% of pages contain reference markers and 32.3\% marginal notes (vs.\ 6.1\% in synthetic data), confirming the high structural density of critical editions, while titles/subtitles occur on 30.7\% of pages. We further classify works into three structural systems based on reference encoding: section systems (21 works), where markers align with block boundaries; milestone systems (7 works), where markers appear within running text; and mixed systems (2 works), combining both. Annotation quality was ensured using a four-eye principle, with each page independently reviewed by two authors; disagreements on \verb|<ref>| and \verb|<note>| were resolved by consulting the full edition. All data sources and code will be released on GitHub upon publication, together with extended metadata and error analyses.

\section{Experimental Design}

\subsection{Data}
We work with two complementary data sources: synthetically rendered page images and real scanned pages. Synthetic images are rendered at high resolution and downsampled to a maximum side length of 1024 pixels; real scans are resized to the same maximum side at inference time. For the synthetic corpus, splits are defined at the level of canonical document groups (i.e., TEI source documents), such that all styled renderings of the same source are assigned to the same split to ensure strict separation of textual content. We allocate 90\% of document groups to training and 10\% to an evaluation pool. The synthetic corpus comprises approximately 185K pages for training, 518 for validation, and 139 for test. The real-data corpus consists of 300 training pages, 60 validation pages, and 90 test pages spanning 30 author-work pairs. Both test splits were constructed to reflect the range of document-level statistics in the evaluation pool, including variation in page length, reference density, marginal note frequency, and heading density.

\subsection{Models and Training Configuration}

We evaluate models spanning general-purpose vision-language models (VLMs) and OCR-specialized architectures selected for their performance in leaderboards: {Qwen3-VL}~\cite{bai2025qwen3vltechnicalreport} (2B and 8B), {DeepSeek-OCR-2}~\cite{wei2026deepseekocr2visualcausal} (3B), and {LightOnOCR-2}~\cite{lightOCR} (1B). Each model is evaluated under four training regimes: (i)~\texttt{zero-shot}; (ii)~\texttt{synthetic}, fine-tuned on the synthetic train set; (iii)~\texttt{real}, fine-tuned directly on the real-data train set; and (iv)~\texttt{synth$\to$real}, fine-tuned first on synthetic data and subsequently continued on real data, initialized from the best checkpoint.

For fine-tuning, we use parameter-efficient LoRA adapters optimized with AdamW. Hyperparameters were selected via grid search over learning rate, LoRA rank, and gradient accumulation steps, with checkpoint selection based on validation CER. 
\finalonly{The final configurations are reported in Table~\ref{tab:hparams}.}
All models use greedy decoding with a maximum of 4096 generated tokens, but since we use different model families, there is a difference in how task instructions are provided. 
LightOnOCR-2 is image-only, without textual prompts, while DeepSeek-OCR-2 uses its built-in \textit{Free OCR} instruction. Qwen3-VL, as a general-purpose VLM, requires an explicit task prompt, which we provide in the form of a structured instruction describing the expected pseudo-Markdown output format. \finalonly{The full prompt is given in Appendix~\ref{fig:qwen_prompt}.}

For baseline OCR performance, we report results under two different settings based on Convolutional Recurrent Neural Networks (CRNN). First, we use Tesseract~\cite{kay2007tesseract}, which requires no additional training and serves as an off-the-shelf baseline. We then use the Kraken software~\cite{kiessling2025version}, first trained on the synthetic dataset and subsequently fine-tuned using the development set only (60 pages). For evaluation, we report results using automatic segmentation with Kraken base segmentation models, reflecting a fully automated pipeline. All Tesseract and Kraken baselines are evaluated exclusively on textual recognition accuracy and do not take document structure into account. 
\finalonly{
  \begin{table}[h]
  \centering
  \small
  \begin{tabular}{llcccccc}
  \toprule
  Model & Phase & LR & Rank & $\alpha$ & Eff.\ batch & Epochs \\
  \midrule
  DeepSeek-OCR-2 & Synthetic       & 2e-4 & 16 & 16 & 32  & 1  \\
                 & Real            & 2e-4 & 32 & 32 & 16   & 20 \\
                 & Synth$\to$Real  & 2e-4 & 32 & 32 & 16   & 20 \\
  \midrule
  LightOnOCR-2   & Synthetic       & 2e-4 & 16 & 32 & 32  & 1  \\
                 & Real            & 6e-5 & 16 & 32 & 16  & 20 \\
                 & Synth$\to$Real  & 6e-5 & 16 & 32 & 16  & 20 \\
  \midrule
  Qwen3-VL       & Synthetic       & 2e-4 & 16 & 16 & 64  & 1  \\
                 & Real            & 2e-4 & 32 & 32 & 8  & 3 \\
                 & Synth$\to$Real  & 2e-4 & 32 & 32 & 8  & 3 \\
  \bottomrule
  \end{tabular}
  \caption{LoRA fine-tuning hyperparameters. Synth$\to$Real warm-starts from
  the best synthetic checkpoint.}
  \label{tab:hparams}
  \end{table}}

\subsection{Evaluation Metrics}
We evaluate transcription quality at two levels: plain-text accuracy and document structure recognition. For the former, we report macro-averaged and median word error rates (WER) and character error rates (CER) computed on raw character sequences after stripping structural markup (markdown and pseudo-XML), applying NFKC normalization, and collapsing whitespace. These metrics are normalized by reference length, allowing values above 100\% if predictions exceed the reference. For structure recognition, we evaluate reference markers (\texttt{<ref>}) and marginal notes (\texttt{<note>}) using longest common subsequence (LCS) F1 in reading order, and header count (\texttt{\#}) using count-based F1. For paragraph indentation markers (\texttt{<tab/>}), we report two accuracy scores: recall (accuracy on paragraphs that should be indented) and specificity (accuracy on paragraphs that should not be indented), measuring whether the model correctly identifies the presence or absence of indentation at the opening paragraph if it could be a continuation from a previous page (i.e., there is no header line before the paragraph). To analyze error patterns, we decompose edit operations at both character and word level into substitutions, insertions, and deletions (reported per 100 reference units). Insertions approximate hallucinations, while deletions capture omitted reference content. Character substitutions are further classified when base Greek letters match, but combining marks differ. We distinguish breathing confusions between \texttt{U+0313} and \texttt{U+0314}, accent confusions (\texttt{U+0300, U+0301, U+0342}), and iota subscript confusions (\texttt{U+0345}).
% If multiple mark types change, the substitution is assigned according to the following priority: breathing > accent > iota subscript. Percentages are computed over total character substitutions, enabling comparison of polytonic error profiles across models.

\section{Results}

Table~\ref{tab:pure_ocr} reports plain-text OCR performance across training regimes and test domains. We observe four main patterns: (i)~baseline models have competitive results OCR-wise, (ii)~fine-tuning substantially improves transcription accuracy over zero-shot inference, (iii)~synthetic supervision transfers effectively to real scanned material, and (iv)~sequential synthetic-to-real training yields the strongest overall performance.

\paragraph{Impact of fine-tuning.}
Fine-tuning consistently improves performance over zero-shot across all models. For instance, Qwen3-VL-8B reduces median CER on real data from 5.2\% (zero-shot) to 2.1\% with real-only training and to 1.0\% under the \texttt{synth$\to$real} regime. Similar gains hold for Qwen3-VL-2B and LightOnOCR-1B, confirming the importance of domain-adapted supervision. Synthetic-only training already transfers strongly to real scans: Qwen3-VL-2B drops from 10.0\% to 2.1\% CER without exposure to real images, and for Qwen3-VL-8B, synthetic-only (1.7\%) even surpasses real-only (2.1\%). Sequential \texttt{synth$\to$real} yields the best results for Qwen models, suggesting complementary effects of broad synthetic coverage and real-domain refinement. This pattern is not universal, however: LightOnOCR-1B performs best with real-only training (1.7\%), while \texttt{synth$\to$real} slightly degrades performance.

\begin{table}[t]
\centering
\caption{Pure OCR results with markup stripped before scoring. Scores are percentages~(\%). Train set indicates fine-tuning data (--- for zero-shot).}
\small
\setlength{\tabcolsep}{4pt}
\renewcommand{\arraystretch}{0.9}
\resizebox{\linewidth}{!}{%
\small
\begin{tabular}{l l l S S S S}
\toprule
Model & Train set & Test set & \multicolumn{1}{c}{CER$_{\text{med}}$} & \multicolumn{1}{c}{CER$_{\text{mean}}$} & \multicolumn{1}{c}{WER$_{\text{med}}$} & \multicolumn{1}{c}{WER$_{\text{mean}}$} \\
\midrule
Tesseract & --- & real & 7.2 & 9.9 & 24.3 & 28.9 \\
% Kraken (Perfect Segmentation) & --- & real & 5.9 & 7.7 & 17.9 & 20.0 \\
Kraken (End to end) & real & real & 5.3 & 6.8 & 16.9 & 9.3 \\
\midrule
\multirow{6}{*}{DeepSeek-OCR-2} 
 & --- & synth & 3.8 & 4.6 & 18.6 & 20.5 \\
 & --- & real & 10.2 & 13.8 & 31.6 & 35.0 \\
 & synth & synth & 1.6 & 98.3 & 7.9 & 111.4 \\
 & synth & real & 5.5 & 52.7 & 15.5 & 71.1 \\
 & real & real & 4.7 & 10.4 & 16.7 & 21.7 \\
 & synth$\to$real & real & 4.7 & 26.7 & 17.0 & 40.7 \\
 \midrule
\multirow{6}{*}{LightOnOCR-1B} 
 & --- & synth & 4.8 & 10.6 & 14.5 & 19.8 \\
 & --- & real & 5.9 & 10.3 & 19.2 & 23.7 \\
 & synth & synth & 4.6 & 12.2 & 11.9 & 19.4 \\
 & synth & real & 5.4 & 14.6 & 18.6 & 25.4 \\
 & real & real & 1.7 & 2.1 & 9.4 & 9.8 \\
 & synth$\rightarrow$real & real & 4.2 & 8.5 & 15.1 & 20.3 \\
 \midrule
\multirow{6}{*}{Qwen3-VL-2B} 
 & --- & synth & 12.4 & 23.2 & 43.1 & 55.8 \\
 & --- & real & 10.0 & 13.1 & 35.4 & 40.5 \\
 & synth & synth & 0.9 & 1.3 & 4.2 & 5.4 \\
 & synth & real & 2.1 & 2.7 & 10.3 & 11.3 \\
 & real & real & 4.0 & 4.7 & 18.1 & 18.2 \\
 & synth$\rightarrow$real & real & 1.5 & 2.2 & 8.2 & 8.8 \\
 \midrule
\multirow{6}{*}{Qwen3-VL-8B}
 & --- & synth & 6.2 & 11.8 & 22.8 & 27.3 \\
 & --- & real & 5.2 & 8.7 & 21.6 & 24.7 \\
 & synth & synth & 0.4 & 0.7 & 2.0 & 3.0 \\
 & synth & real & 1.7 & 1.9 & 8.4 & 9.2 \\
 & real & real & 2.1 & 2.3 & 10.8 & 10.9 \\
 & synth$\rightarrow$real & real & \textbf{1.0} & \textbf{1.2} & \textbf{5.6} & \textbf{6.6} \\
\bottomrule
\end{tabular}
}
\label{tab:pure_ocr}
\end{table}

\paragraph{Model stability and error dispersion.}
Several settings exhibit substantial gaps between median and mean error rates, particularly for DeepSeek-OCR-2. For instance, under synthetic training evaluated on synthetic data, the model attains a low median CER (1.6\%) but an extremely high mean CER (98.3\%), indicating catastrophic page-level failures. This divergence suggests that while many pages are transcribed accurately, a subset produces very long or unstable outputs that inflate mean error rates. In contrast, LightOnOCR-1B shows much smaller median-mean discrepancies, indicating more stable behavior. Among all models, Qwen3-VL-8B achieves the strongest performance, reaching a median CER of 1.0\% and median WER of 5.6\% under the \texttt{synth$\to$real} regime on real scanned pages. These results demonstrate that large vision-language models, when combined with structure-aware synthetic supervision and real-domain adaptation, can achieve near-perfect plain-text OCR on complex critical editions.

\paragraph{CRNNs remain competitive.} 
An examination of the results, particularly under zero-shot conditions on the real dataset, shows that Tesseract outperforms two out of four models in terms of median CER and three out of four in terms of mean CER.\footnote{It should also be noted that the WER and CER scores of both CRNN-based systems are inherently affected by the overprediction of headers and line numbers. These elements are ignored in ground truth transcriptions for the end-to-end pipelines and therefore introduce systematic noise into the evaluation metrics.} In both cases, Qwen3-VL-8B is the only model that consistently maintains superior performance. Under similar conditions using only real training examples, Kraken achieves better mean CER than DeepSeek, although it is otherwise outperformed within relatively modest margins, especially when considering the respective computational footprints of these models, except for LightOnOCR and Qwen3 models in a \texttt{synth$\rightarrow$real} fine-tuning scenarios.

\begin{table}[t]
\centering
\caption{Structure recognition (\%) on the \texttt{real} test set. Headers: count-based F1; references and notes: longest common subsequence F1. Tab$_{1\text{st}}$: specificity (Spec) and recall (Rec) for the first paragraph with indentation.}% Train \texttt{---} denotes zero-shot.}
\small
\setlength{\tabcolsep}{3pt}
\renewcommand{\arraystretch}{0.9}
\resizebox{\linewidth}{!}{%
\small
\begin{tabular}{l l S[table-format=3.1] S[table-format=3.1] S[table-format=3.1] S[table-format=3.1] S[table-format=3.1]}
\toprule
Model & Train & \multicolumn{1}{c}{Hdr F1} & \multicolumn{1}{c}{Ref F1} & \multicolumn{1}{c}{Note F1} & \multicolumn{1}{c}{Tab$_{1\text{st}}$ Spec} & \multicolumn{1}{c}{Tab$_{1\text{st}}$ Rec} \\
\midrule
\multirow{3}{*}{\textsc{DeepSeek-OCR-2}} & \texttt{synth} & 18.9 & 47.6 & 0.0 & 100.0 & 0.0 \\
 & \texttt{real} & 27.1 & 46.0 & 13.9 & 100.0 & 0.0 \\
 & \texttt{synth$\to$real} & 39.6 & 56.7 & 27.2 & 100.0 & 0.0 \\
\addlinespace[2pt]
\hline
\addlinespace[2pt]
\multirow{3}{*}{\textsc{LightOnOCR-1B}} & \texttt{synth} & 39.0 & 43.9 & 7.4 & 100.0 & 0.0 \\
 & \texttt{real} & 27.1 & 72.8 & 47.3 & 92.3 & 38.5 \\
 & \texttt{synth$\to$real} & 21.9 & 51.5 & 19.9 & 100.0 & 0.0 \\
\addlinespace[2pt]
\hline
\addlinespace[2pt]
\multirow{4}{*}{\textsc{Qwen3-VL-2B}} & \multicolumn{1}{c}{---} & 0.0 & 3.3 & 0.0 & 42.0 & 53.3 \\
 & \texttt{synth} & 41.1 & 59.2 & 0.0 & 78.2 & 33.3 \\
 & \texttt{real} & 55.9 & 65.4 & 11.8 & 81.1 & 78.6 \\
 & \texttt{synth$\to$real} & 75.8 & 76.2 & 39.1 & 92.8 & 80.0 \\
\addlinespace[2pt]
\hline
\addlinespace[2pt]
\multirow{4}{*}{\textsc{Qwen3-VL-8B}} & \multicolumn{1}{c}{---} & 15.3 & 15.7 & 0.0 & 62.3 & 40.0 \\
 & \texttt{synth} & 70.3 & 63.2 & 6.3 & 88.4 & 53.3 \\
 & \texttt{real} & 62.0 & 71.9 & 14.5 & 88.4 & 66.7 \\
 & \texttt{synth$\to$real} & 80.1 & 79.5 & 63.5 & 88.4 & 93.3 \\
\bottomrule
\end{tabular}
}
\label{tab:struct_real}
\end{table}

\paragraph{Structural element recognition.} Table \ref{tab:struct_real} reports structure recognition performance on the real test set. Reference markers achieve the highest F1 scores across all models and training regimes, with Qwen3-VL-8B (synth$\rightarrow$real) reaching 79.5. Marginal note detection is the most training-sensitive element: synth-only fine-tuning yields near-zero Note F1 for Qwen models (0.0-6.3), consistent with their underrepresentation in the synthetic corpus, while real-domain exposure recovers performance substantially (up to 63.5 for Qwen3-VL-8B synth$\rightarrow$real). Header generation remains weak across all models, with only the two Qwen models achieving a count-based F1 score above 75.0 (best: 80.1 for Qwen3-VL-8B synth$\rightarrow$real), likely due to the high variability in heading styles across editions. Tab indentation separates instruction-tuned from OCR-specialized models: Qwen models show calibrated specificity/recall trade-offs, while DeepSeek and LightOnOCR almost never predict indentation, yielding trivially perfect specificity at zero recall. Overall, Qwen3-VL-8B under the synth$\rightarrow$real regime achieves the strongest structure recognition performance across all element types.

\begin{table}[htbp]
\centering
\caption{%
Edit rates and diacritic breakdown on real data.
\textit{cSub/cIns/cDel}: per 100 chars; 
\textit{wSub/wIns/wDel}: per 100 words.
\textit{Br./Ac./Is.}: breathing, accent, iota subscript confusions (\% of cSub).
% \textbf{Bold} = best per column.%
}
\label{tab:errors-real}
\renewcommand{\arraystretch}{0.9}
\setlength{\tabcolsep}{5pt}
\small
\resizebox{\linewidth}{!}{%
\begin{tabular}{l
    S[table-format=2.2]
    S[table-format=2.2]
    S[table-format=2.2]
    !{\vrule width 0.4pt}
    S[table-format=2.2]
    S[table-format=2.2]
    S[table-format=2.2]
    !{\vrule width 0.4pt}
    S[table-format=2.1]
    S[table-format=2.1]
    S[table-format=2.1]
  }
\toprule
 &
  \multicolumn{3}{c}{\textsc{Character edits}} &
  \multicolumn{3}{c}{\textsc{Word edits}} &
  \multicolumn{3}{c}{\textsc{Diacritics}} \\
\cmidrule(lr){2-4}\cmidrule(lr){5-7}\cmidrule(lr){8-10}
Model &
  {\textit{cSub}} & {\textit{cIns}} & {\textit{cDel}} &
  {\textit{wSub}} & {\textit{wIns}} & {\textit{wDel}} &
  {\textit{Br}} & {\textit{Ac}} & {\textit{Is}} \\
\midrule
\textsc{DeepSeek-OCR-2}     & 8.94          & 3.15          & 1.10          & 44.21          & 3.85          & 1.99          & 4.8           & 18.2          & 1.5          \\
\textsc{LightOnOCR-1B}       & 6.10          & 5.73          & 1.11          & 35.22          & 6.87          & 2.94          & 7.2           & 6.6           & 1.5          \\
\textsc{Qwen3-VL-2B}         & 10.12         & 3.22          & 1.86          & 52.22          & 4.69          & 3.73          & 11.3          & 24.4          & 1.6          \\
\textsc{Qwen3-VL-8B}        & 7.02          & 3.58          & 2.12          & 39.61          & 2.83          & 2.94          & 7.3           & 15.7          & 1.3          \\
\midrule
\textsc{DeepSeek-OCR-2}\textsubscript{\texttt{real}}   & \textbf{3.22} & 4.57          & 2.10          & \textbf{18.47} & 5.18          & 3.17          & 9.9           & 17.1          & 2.4          \\
% \textsc{DeepSeek-OCR-2}\textsubscript{\texttt{synth}}  & 5.09          & 7.14          & 7.22          & 29.96          & 12.38         & 8.66          & \textbf{0.7}  & 3.1           & 0.3          \\
\textsc{LightOnOCR-1B}\textsubscript{\texttt{real}} & 5.61          & 0.44          & 0.60          & 33.09          & 1.06          & 1.49          & 5.2           & 3.3           & 1.2          \\
% \textsc{LightOnOCR}\textsubscript{\texttt{synth}} & 8.18          & 1.14          & 6.56          & 35.10          & 2.87          & 9.75          & 2.4           & 2.7           & 0.4          \\
\textsc{Qwen3-VL-2B}\textsubscript{\texttt{real}}    & 6.81          & 0.71          & 0.84          & 36.95          & 1.58          & 1.98          & 6.3           & 12.8          & 1.0          \\
% \textsc{Qwen3-VL-2B}\textsubscript{\texttt{synth}}  & 5.46          & 0.36          & 0.53          & 32.34          & 0.83          & 2.72          & 1.9           & 3.9           & 0.4          \\
\textsc{Qwen3-VL-8B}\textsubscript{\texttt{real}}    & 5.81          & 0.33 & 0.46 & 33.42          & 0.86          & \textbf{1.20} & 4.0           & 7.2           & 1.0          \\
% \textsc{Qwen3-VL-8B}\textsubscript{\texttt{synth}}   & 5.35          & 0.35          & 0.51          & 32.03          & \textbf{0.72} & 1.95          & 1.3           & \textbf{2.6}  & \textbf{0.2} \\
\textsc{DeepSeek-OCR-2}\textsubscript{\texttt{synth$\to$real}} & 5.71 & 11.64 & 6.59 & 32.26 & 15.59 & 7.72 & 3.9 & 7.2 & 1.1 \\
\textsc{LightOnOCR-1B}\textsubscript{\texttt{synth$\to$real}} & 7.72 & 1.08 & 1.79 & 35.06 & 2.35 & 5.61 & 2.1 & 3.0 & \textbf{0.3} \\
\textsc{Qwen3-VL-8B}\textsubscript{\texttt{synth$\to$real}}  & 5.28 & 0.23 & \textbf{0.33} & 31.34 & 0.71 & 2.12 & 1.7 & 3.3 & 0.4 \\
\textsc{Qwen3-VL-8B}\textsubscript{\texttt{synth$\to$real}}  & 5.05 & \textbf{0.22} & 0.35 & 30.39 & \textbf{0.68} & 1.43 & \textbf{1.1} & \textbf{2.1} & \textbf{0.3} \\
\bottomrule
\end{tabular}
}
\end{table}

\paragraph{Error pattern analysis.}
Zero-shot models exhibit high substitution rates at both character (6.10--10.12 per 100 chars) and word level (35--52 per 100 words), indicating substantial
lexical instability. Fine-tuning consistently reduces substitutions: \textsc{DeepSeek-OCR-2}\textsubscript{\texttt{real}} achieves the lowest cSub (3.22) and wSub (18.47), roughly halving its zero-shot error. Real fine-tuning
also nearly eliminates insertions for \textsc{Qwen} and \textsc{LightOnOCR} models (down to 0.22--0.71), suggesting improved alignment with page content,
although \textsc{DeepSeek-OCR-2}\textsubscript{\texttt{real}} remains insertion-prone. Sequential \texttt{synth$\to$real} training further reduces insertion and deletion rates for \textsc{Qwen3-VL-8B} (cIns = 0.22, cDel = 0.35), while also yielding the lowest diacritic confusion rates overall. Manual inspection shows that errors in the best-performing models are primarily orthographic and normalization-related. Typical issues include hyphenation inconsistencies, omitted macrons or breathings, folio insertion artifacts (e.g., embedded page numbers), and occasional Latin look-alike substitutions. Overall, textual grounding remains stable, with errors concentrated at the character level. In contrast, the worst-performing models display elevated substitution and insertion rates driven largely by formatting intrusions and decoding artifacts. These include spurious HTML tags, stray \LaTeX{} fragments, unintended Latin characters, and hallucinated Ancient Greek text. Such artifacts introduce non-source tokens and fragment otherwise correct spans, and often result in length overgeneration, suggesting generative drift during decoding rather than simple orthographic noise.

\section{Discussion}

\paragraph{Script mixing, orthographic instability and symbolic ambiguity.}
Across models, we observe systematic confusion between visually similar characters across scripts, including Greek and Latin capitals (e.g., Greek Alpha vs. Latin A). In zero-shot conditions, biases from scientific pretraining corpora occasionally bleed into the output, with \LaTeX{} formatting used to transcribe Greek script. Such substitutions are particularly problematic in critical editions, where case and script distinctions may encode structural or semantic information. More broadly, these errors reflect a deeper symbolic ambiguity: the same glyph may function as a lexical unit, a numeral, or part of a reference marker depending on context. In Greek critical editions, alphabetic numerals reuse letterforms as numeric symbols, and similar ambiguities arise in marginal line numbers and hierarchical markers. When contextual language priors dominate visual evidence, models fail to reliably disambiguate these multifunctional symbols, and errors often propagate to structure recognition. This suggests that structure-aware OCR for historical material requires more explicit modeling of document-level symbolic systems rather than relying solely on token-level language modeling.

\paragraph{Instability and catastrophic failures.}
The divergence between median and mean error rates, especially for DeepSeek-OCR-2, reveals that some models produce catastrophic page-level failures despite strong median performance. These cases are often characterized by long, fluent outputs that deviate from the visual input, suggesting over-reliance on internal language priors. While large VLMs can achieve near-perfect performance on many pages, their behavior remains brittle in structurally dense or symbol-heavy contexts.

\paragraph{Implications for document understanding.} Beyond Ancient Greek critical editions, our findings carry broader implications for document understanding research. High plain-text accuracy does not guarantee structural competence: models with near-perfect character error rates may still fail to recover hierarchical markers, marginal annotations, or reference systems. Classical OCR baselines remain competitive in transcription while avoiding the catastrophic generative failures observed in some vision–language models, highlighting that accuracy and structural reliability are distinct dimensions. Moreover, the divergence between median and mean error rates shows that generative vision–language models can exhibit unstable behavior, producing fluent but visually ungrounded outputs. Overall, our results suggest that document understanding in historical material requires modeling not only glyph recognition but also the hierarchical and symbolic systems encoded in layout. Structure-aware OCR is therefore a problem of interpretation, not merely transcription.
\section{Conclusion}

In this paper, we introduced a new document type for joint text and layout recognition, focusing on critical editions of polytonic Greek texts, together with a reusable open-source pipeline and two complementary datasets (synthetic and manually annotated scans). Our results show that most models struggle out-of-the-box on this domain, and that while some vision--language models can adapt to both script and structure, they remain prone to hallucinations that we were not able to fully mitigate. A key takeaway of our study is that classical OCR pipelines remain competitive in terms of efficiency and stability for text recognition. While vision--language models provide clear advantages in adaptability and structure awareness, particularly when trained sequentially on synthetic and real data, these gains come at a substantially higher computational cost. VLM inference requires GPU execution (e.g., 0.09--0.16 kWh per run on A100-class hardware), whereas classical OCR systems operate on CPU with significantly lower energy and runtime requirements. Overall, our findings suggest that vision--language models are powerful but not universally superior solutions for structure-aware OCR. Rather than replacing classical pipelines, they offer complementary strengths. Future work should therefore explore hybrid approaches that combine the structural sensitivity of VLMs with the efficiency and robustness of traditional OCR systems.

\finalonly{
\section*{Acknowledgments}

The project ``Corpus Liberatum Linguae Graecae'' was supported by the French National Research Agency (ANR) under the France 2030 grant reference number ``ANR-24-RRII-0002'' operated by the Inria Quadrant Program and is endorsed by the BPI Scribe project.
}

\section*{Appendix}

\begin{tcolorbox}[
  title=Annotation Guidelines,
  colback=gray!5,
  colframe=black,
  fonttitle=\bfseries,
  boxrule=0.7pt,
  fontupper=\footnotesize,
  fontlower=\small
]

\textbf{Scope.}  
Annotations capture \emph{only} the main textual content, the structural elements and marginal notes. 
Running headers, running footers, page numbers, and other layout artifacts are explicitly excluded.

\vspace{0.2em}
\textbf{Structure.}  
We annotate three types of Markdown-style headings:

- \verb|# TITLE| --- section title without an explicit reference marker.

- \verb|# <ref>REF</ref>| --- section title consisting solely of a reference marker.

- \verb|# <ref>REF</ref> TITLE| --- section title with both a reference and a title.

\vspace{0.2em}
\textbf{Inline annotation elements.}  
Inline annotations are restricted to the following two elements:

- \verb|<ref>...</ref>| --- reference or milestone markers embedded in running text.

- \verb|<note>...</note>| --- marginal notes, glosses, or editorial comments.

- We add \verb|<tab/>| to paragraphs that start on a page with indetation.
%Inline annotations do not introduce new structural blocks and %preserve the surrounding text verbatim.
\label{app:annotation-guidelines}
\end{tcolorbox}

\finalonly{\begin{figure*}[h]
\centering
\footnotesize
\fbox{
\begin{minipage}{0.95\textwidth}
\textbf{Prompt used for Qwen3-VL.}

\medskip

\textit{You are an OCR system for Ancient Greek and Latin printed scholarly texts. 
Transcribe the page exactly in pseudo-Markdown.}

\medskip

\textbf{Rules:}
Copy text exactly (retain all characters, accents, ligatures, punctuation, spacing). Do not normalize, correct, or translate. Follow visual reading order. Each paragraph must be output as one line. Add \texttt{<tab/>} at the start of paragraphs that clearly begin on this page; do not add it if the paragraph continues from a previous page. Join words split by line-break hyphens. Wrap section markers as \texttt{<ref>X</ref>}. Major headers use:
\texttt{\# <ref>X</ref>} or 
\texttt{\# <ref>X</ref> TITLE}. 
Standalone titles use \texttt{\# TITLE}. 
Inline and marginal section markers are encoded inline. Margin notes are encoded as \texttt{<note>TEXT</note>}. Ignore running heads, footers, page numbers, line numbers, printer marks, and footnote markers. Output only the transcription.
\end{minipage}
}
\caption{Instruction prompt provided to Qwen3-VL for structure-aware OCR transcription.}
\label{fig:qwen_prompt}
\end{figure*}}

\finalonly{%
\begin{figure}[htbp]
\centering

\begin{minipage}{0.49\textwidth}
\centering
\includegraphics[width=\linewidth]{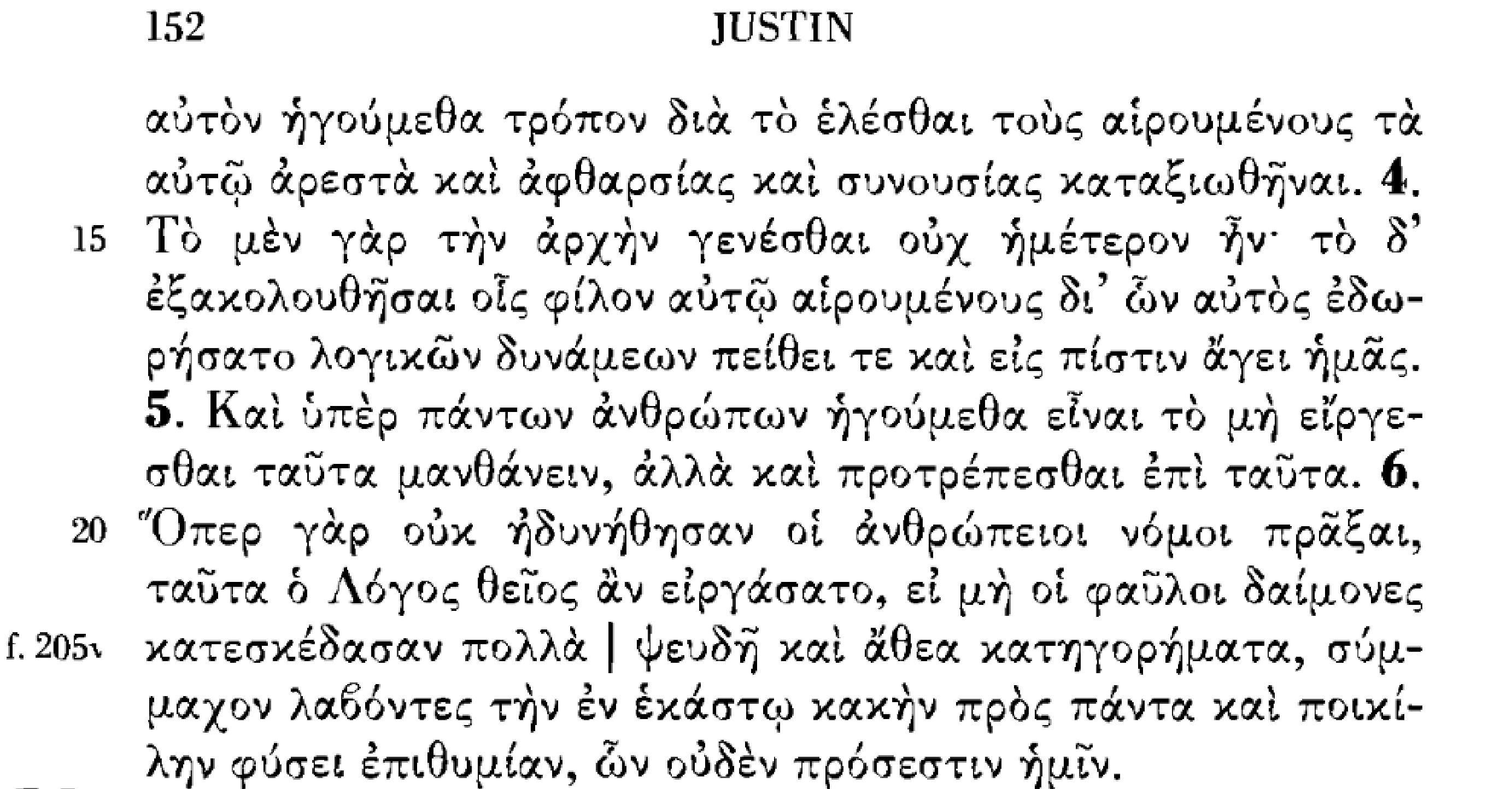}
\end{minipage}
\hfill
\begin{minipage}{0.49\textwidth}

\begin{greek}
    {\tiny 
αὐτὸν ἡγούμεθα τρόπον διὰ τὸ ἑλέσθαι τοὺς αἱρουμένους τὰ\\[-1em]
αὐτῷ ἀρεστὰ καὶ ἀφθαρσίας καὶ συνουσίας καταξιωθῆναι. <ref>4</ref>\\[-1em]
Τὸ μὲν γὰρ τὴν ἀρχὴν γενέσθαι οὐχ ἡμέτερον ἦν· τὸ δ\DIFdelbegin \DIFdel{'}\DIFdelend \DIFaddbegin \DIFadd{’}\DIFaddend\\[-1em]
ἐξακολουθῆσαι οἷς φίλον αὐτῷ αἱρουμένους δι\DIFdelbegin \DIFdel{' }\DIFdelend \DIFaddbegin \DIFadd{’ }\DIFaddend ὧν αὐτὸς \DIFdelbegin \DIFdel{ἐδωρήσατο}\DIFdelend \DIFaddbegin \DIFadd{ἐδω‐\\[-1em] ρήσατο }\DIFaddend 
λογικῶν δυνάμεων πείθει τε καὶ εἰς πίστιν ἄγει ἡμᾶς.\\[-1em]
<ref>5</ref> Καὶ ὑπὲρ πάντων ἀνθρώπων \\[-1em]\hspace*{5em}ἡγούμεθα εἶναι τὸ μὴ \DIFdelbegin \DIFdel{εἴργεσθαι}\DIFdelend \DIFaddbegin \DIFadd{εἴργε‐\\[-1em] σθαι }\DIFaddend
ταῦτα μανθάνειν, ἀλλὰ καὶ προτρέπεσθαι ἐπὶ ταῦτα. <ref>6</ref>\\[-1em]
Ὅπερ γὰρ οὐκ ἠδυνήθησαν οἱ ἀνθρώπειοι νόμοι πρᾶξαι,\\[-1em]
ταῦτα ὁ Λόγος θεῖος ἂν εἰργάσατο, εἰ μὴ οἱ φαῦλοι δαίμονες\\[-1em]
\DIFdelbegin \DIFdel{<note>f. 205v</note> }\DIFdelend κατεσκέδασαν πολλὰ | ψευδῆ καὶ\\[-1em]\hspace*{5em} ἄθεα κατηγορήματα, \DIFdelbegin \DIFdel{σύμμαχον}\DIFdelend \DIFaddbegin \DIFadd{σύμ‐ \\[-1em]μαχον }\DIFaddend 
λαβόντες τὴν ἐν ἑκάστῳ κακὴν πρὸς πάντα καὶ \DIFdelbegin \DIFdel{ποικίλην}\DIFdelend \DIFaddbegin \DIFadd{ποικί‐ λην}\DIFaddend\\[-1em]
φύσει ἐπιθυμίαν, ὧν οὐδὲν πρόσεστιν ἡμῖν.}
\end{greek}
\end{minipage}
\caption{Example of output of Qwen3.5-VL.8b on the test set.}
\end{figure}
}

\bibliographystyle{splncs04}
\bibliography{bibliography}

\end{document}